\documentclass{article}
\usepackage{ijcai19}

\usepackage{times}%
\usepackage{url}
\usepackage[hidelinks]{hyperref}
\usepackage[utf8]{inputenc}
\usepackage[small]{caption}
\usepackage{graphicx}%
\usepackage{amsmath}%
\usepackage{booktabs}
\usepackage{algorithm}
\usepackage{epsfig}%
\usepackage{caption}%
\usepackage{amssymb}%
\usepackage{subfigure}%
\usepackage{algorithmic}
\title{Resolution-invariant Person Re-Identification}

\author{
Shunan Mao$^1$
\and
Shiliang Zhang$^1$\And
Ming Yang$^{2}$
\affiliations
$^1$Peking University\\
$^2$Horizon Robotics, Inc
\emails
\{snmao, slzhang.jdl\}@pku.edu.cn,
ming.yang@horizon-robotics.com
}

\begin{document}

\maketitle

\begin{abstract}
Exploiting resolution invariant representation is critical for person Re-Identification (ReID) in real applications, where the resolutions of captured person images may vary dramatically. This paper learns person representations robust to resolution variance through jointly training a Foreground-Focus Super-Resolution (FFSR) module and a Resolution-Invariant Feature Extractor (RIFE) by end-to-end CNN learning. FFSR upscales the person foreground using a fully convolutional auto-encoder with skip connections learned with a foreground focus training loss. RIFE adopts two feature extraction streams weighted by a dual-attention block to learn features for low and high resolution images, respectively. These two complementary modules are jointly trained, leading to a strong resolution invariant representation. We evaluate our methods on five datasets containing person images at a large range of resolutions, where our methods show substantial superiority to existing solutions. For instance, we achieve Rank-1 accuracy of 36.4\% and 73.3\% on \emph{CAVIAR} and \emph{MLR-CUHK03}, outperforming the state-of-the art by 2.9\% and 2.6\%, respectively.
\end{abstract}

\section{Introduction}

\noindent Person Re-identification (ReID) aims to find a probe person from a large-scale person image gallery collected by a camera network.~\cite{li2019multi}
Person ReID is challenging since it is confronted by many appearance variations due to camera viewpoint, person pose, illumination, background, \emph{etc}. Thanks to the introduction of many benchmark datasets like \emph{VIPeR}~\cite{gray2008viewpoint}, \emph{CUHK03}~\cite{li2014deepreid}, \emph{Market1501}~\cite{zheng2015scalable} and \emph{MSMT17}~\cite{wei2018person}, most of these challenges are covered in these datasets, leading to a significant progress in person ReID performance.

\begin{figure}
    \centering
    \includegraphics[width=1\linewidth]{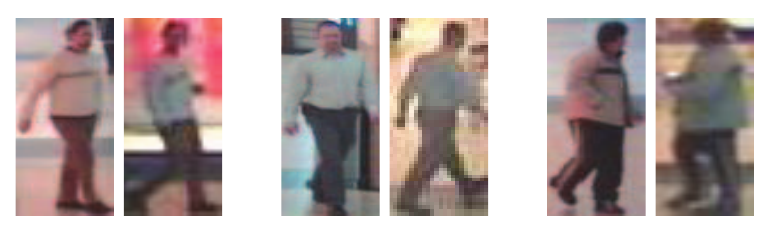}\\
    \caption{Illustration of 6 images from 3 persons in \emph{CAVIAR} [Cheng \emph{et al}., 2011]. Person ReID needs to match the same person and discern different persons across different resolutions.}
    \vspace {-3mm}
    \label{fig:difR}
\end{figure}

Among the above challenges, varying resolutions of person images are probably the most common one, due to the distance to a camera, or camera focus and resolution. Matching persons at different resolutions requires the ReID algorithms to attend to distinct visual cues. For example, Fig.~\ref{fig:difR} illustrates two instances of three persons sampled from \emph{CAVIAR}~\cite{cheng2011custom}. With high resolution image samples, those three persons can be distinguished by their hair styles or strips on the pants. As these details are not available in low resolution images, a ReID method needs to resort to silhouettes or global textures for a reliable matching. Moreover, the high and low resolution samples of the same person may even present a larger discrepancy than to those samples from different persons at a similar resolution. Therefore, dedicated treatments are desired for ReID methods to cope with large resolution variations of person images.

Matching persons at dramatically different resolutions has not been extensively studied,
partly because of the limitation of current ReID benchmark datasets.
Most widely used benchmark datasets usually consist of person images with limited resolution variations.
\emph{CAVIAR}~\cite{cheng2011custom} is particularly collected to consider two levels of resolutions.
\emph{MLR-VIPeR} and \emph{MLR-CUHK03}~\cite{jiao2018deep} are adapted from \emph{VIPeR}~\cite{gray2008viewpoint} and \emph{CUHK03}~\cite{li2014deepreid} by including three levels of resolutions, respectively.
These datasets have inspired many works on low-resolution person ReID~\cite{li2015multi,jing2015super,wang2016scale,jiao2018deep,wang2018cascaded},
yet not many efforts on how to handle person images with a large range of resolution variance.

Traditional methods~\cite{li2015multi,jing2015super,wang2016scale} address person ReID with varying person resolutions mainly by learning a shared feature space between low and high resolutions. Recent approaches focus on deep learning based Super-Resolution (SR)~\cite{jiao2018deep,wang2018cascaded}.
Although SR methods can recover some visual details, they do not differentiate person foregrounds and backgrounds and are not optimized for person ReID, \emph{i.e.}, their goal is to minimize the pixel-level L2 loss, rather than the person ReID errors. In practice, SR methods are not capable to fully recover the missing details in low resolution images. We argue that the person feature extractor shall be explicitly designed and optimized to combat against challenging resolution variance in real-world scenarios.

This paper proposes to jointly optimize person image resolution and feature extraction for person ReID. Specifically, we propose a deep network consisting of two modules.
The Foreground-Focus Super-Resolution (FFSR) module upscales the resolution of an input image using a fully convolutional auto-encoder with skip connections.
Different from general SR modules, FFSR is jointly trained with the person ReID loss and a foreground focus loss, which recovers the details on the person body and suppresses the cluttered backgrounds.
The subsequent Resolution-Invariant Feature Extractor (RIFE) extracts person representations for person ReID. RIFE consists of several feature learning blocks, each of which adopts two CNN branches to learn features from low and high-resolution images, respectively. This design learns more dedicated feature extractors for low-resolution inputs. In other words, RIFE explicitly differentiates high and low resolution inputs during feature learning to ensure its robustness to resolution variance. Features from those two branches are fused with the weights predicted by a Dual-Stream Block (DSB) as the resolution invariant feature.

By jointly training FFSR and RIFE, our approach achieves consistent improvements on the three multi-resolution ReID datasets, \emph{i.e.}, \emph{CAVIAR}~\cite{cheng2011custom}, \emph{MLR-VIPeR}, and \emph{MLR-CUHK03}~\cite{jiao2018deep}. Besides those three datasets, we also construct two large datasets with large variations of person resolutions, \emph{i.e.}, \emph{VR-Market1501} and \emph{VR-MSMT17} by modifying \emph{Market1501}~\cite{zheng2015scalable} and \emph{MSMT17}~\cite{wei2018person}, respectively. On these two datasets, our method also achieves promising performance. To our best knowledge, this is an original work that jointly considers foreground focus super resolution and multiple CNN branches for resolution invariant representations in person ReID. Extensive ablation studies as well as comparisons on five datasets have shown the competitive performance of the proposed approach.

\section{Related Work}
This section briefly reviews low-resolution person ReID and image super-resolution, which are closely related to our work.

\vspace {3mm}

\noindent\textbf{Low-Resolution Person Re-ID}. Some works use matric learning methods to address low-resolution person ReID mainly by learning a shared feature space between low and high resolutions. For example, JUDEA~\cite{li2015multi} optimizes the distance between images of different resolutions by requiring features on the same person to be close to each other. SLD$^2$L~\cite{jing2015super} uses the Semi-Coupled Low-Rank dictionary learning to build the mapping between features from low and high-resolution images. SDF~\cite{wang2016scale} learns a discriminating surface to separate feasible and infeasible functions in the scale distance function space.
\cite{yu2019learning} uses adversarial loss and reconstruction loss to decrease distance between deep features from different resolution.
Other works use deep learning based Super-Resolution (SR). CSR-GAN~\cite{wang2018cascaded} focuses on the super resolution part, and uses a deep Cascaded SR-GAN as well as several handcraft restrictions to enhance the image resolution. SING~\cite{jiao2018deep} adds a Super Resolution network before the feature extraction and trains two networks jointly.

\vspace {3mm}

\noindent\textbf{Super-Resolution} benefits from the advance of deep models. SRCNN~\cite{dong2014learning} first introduces a Fully Convolutional Network for image Super-Resolution. Many works~\cite{kim2016accurate,tai2017image} have been proposed by designing deeper, wider, and denser network architectures. SRGAN~\cite{ledig2017photo} designs additional loss functions to recover more semantic cues. Those works are general SR models and do not concern with image contents. SFT-GAN~\cite{wang2018recovering} uses image segmentation to help the texture super resolution.~\cite{yu2018super} use an encoder-decoder structure to leverage attributes and use GAN~\cite{goodfellow2014generative} and STN~\cite{jaderberg2015spatial} to make the generated faces appear realistic.

\vspace {3mm}

Different from SING and CSR-GAN, our FFSR focuses on person foreground and RIFE learns different feature extractors for high and low-resolution images. This further enhances the robustness to resolution variance.
\section{Problem Formulation}
In surveillance videos, a person image can be regarded as a sample of one person captured by a camera, where the resolution is decided by shooting parameters like sensor resolution, shooting distance, camera focus, imaging processor, \emph{etc}. \emph{i.e.},
\begin{equation}\label{eq:imgsamp}
    I_{i}^{r}= \operatorname{camera-sample}(P_{i}, \theta),
\end{equation}
where $I_{i}^{r}$ is a person image with resolution $r$ and image index $i$. $P_{i}$ denotes the person ID label of $I_{i}^{r}$, $\theta$ denotes the shooting parameters.

It is hard to precisely define the resolution $r$, because the parameters $\theta$ could be complicated. For simplicity, for $I_{i}^{r}$ in a dataset $\mathbb D$, we use a scalar $r\in[0,1]$, computed with $width(I_{i}^{r})/width_{max}$ as its resolution, where $width_{max}$ is the maximum width of images on $\mathbb D$. For example with $width_{max}=96$, resizing an original $128\times48$ sized image to $64\times24$ degrades its resolution from 0.5 to 0.25. To simplify the definition of resolution, we note that enlarging an image with interpolation does not enhance its resolution.

The task of person ReID can be described as matching a query person against the collected person image dataset using a feature representation $f$, with the goal of minimizing the distance between images of the same person, meanwhile maintaining larger distances between images of different persons. Considering the variance of image resolution, we denote the objective function $\mathcal {O}$ of person ReID as,
\begin{equation}\label{eq:obj}
\begin{aligned}
\mathop{\min}_{f}\ \mathcal{O}(r_1, r_2) & ={\operatorname D_{sim}({r_{1},r_{2}})}/{\operatorname D_{dif}({r_{1},r_{2}})}, \\
 \ \  \operatorname D_{sim}(r_{1},r_{2}) & \triangleq\sum_{P_i=P_{i'}}^{}\|f_{i}^{r_{1}}-f_{i'}^{r_{2}}\|_{2}^{2}, \\
\operatorname D_{dif}({r_{1},r_{2}}) & \triangleq\sum_{P_i\neq P_{i'}}^{}\|f_{i}^{r_{1}}-f_{i'}^{r_{2}}\|_{2}^{2}, \\
\end{aligned}
\end{equation}
where $\|\cdot\|_{2}^{2}$ computes the distance between feature vectors. $\operatorname D_{dif}(\cdot)$ and $\operatorname D_{sim}(\cdot)$ compute the distance between two images of the same person and different persons, respectively. We use superscripts $r_1$ and $r_2$ to denote resolutions of two images considered in distance computation.

\begin{figure}
    \centering
    \includegraphics[width=0.9\linewidth]{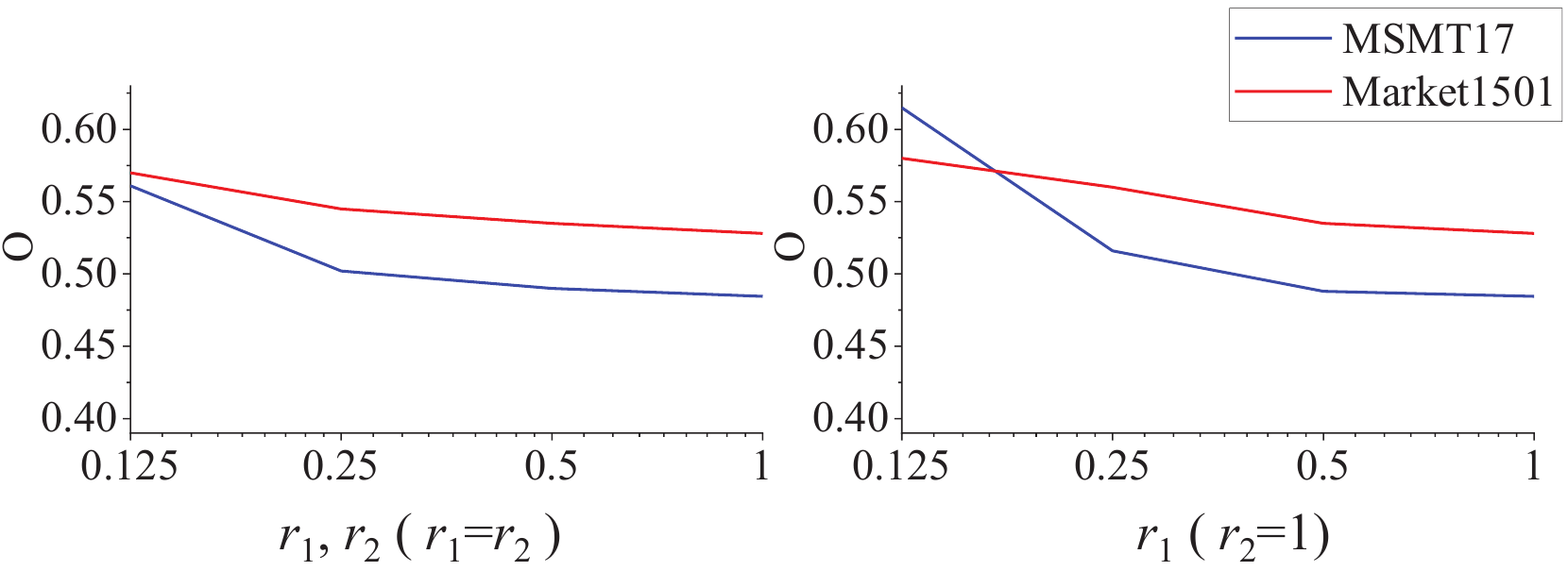}\\
    \caption{Values of object function $\mathcal{O}$ in Eq.~\eqref{eq:obj} computed with variations of resolution on \emph{MSMT17} and \emph{Market1501}. (a) fixes $r_1 = r_2$ and increase $r_1$ and $r_2$ from 0.125 to 1. (b) fixes $r_2=1$ and increase $r_1$ from 0.125 to 1. It verifies that, both low resolution and varied resolution increase the difficulty of person ReID.}
    \label{fig:SwSb}
\end{figure}

Before proceeding to the formulation of our algorithm, we first illustrate the effects of resolution variance to person ReID performance on two large ReID datasets~\emph{Market1501}~\cite{zheng2015scalable} and~\emph{MSMT17}~\cite{wei2018person}, respectively. We first train a ResNet50 baseline~\cite{he2016deep} as the feature extractor, then compute $\mathcal{O}(r_1, r_2)$ on two datasets with different combinations of $r_1$ and $r_2$. Fig.~\ref{fig:SwSb} (a) fixes $r_1 = r_2$ and increases their values from 0.125 to 1. We observe that, lower resolution leads to larger $\mathcal{O}$, resulting in a lower person ReID accuracy. Fig.~\ref{fig:SwSb} (b) fixes $r_2 =1$ and increases $r_1$ from 0.125 to 1. It is clear that, larger variance of resolution corresponds to increased person ReID difficulty. We also observe that, the curves in Fig.~\ref{fig:SwSb} (b) are more abrupt than the ones in Fig.~\ref{fig:SwSb} (a), indicating that varied-resolution ReID could be more challenging than the low-resolution case.

Our solution is inspired by the above observations, \emph{i.e.}, to improve person ReID accuracy, two compared images should present at 1) high resolution and 2) similar resolution. The person image resolution should be enhanced to recover visual details. To facilitate feature extraction, the SR model is expected to focus on the person foreground and suppress the cluttered backgrounds. Meanwhile, the feature extractor should be able to alleviate the resolution variances. Those two intuitions correspond to two modules in our network, \emph{i.e.}, the Foreground Focus Super Resolution (FFSR) and Resolution Invariant Feature Extractor (RIFE), respectively.

For an input person image $I_{i}^{r}$, FFSR first enhances its resolution to $r',r \leq r'$, then it is processed by RIFT for resolution invariant feature extraction. The forward computation of our network can be denoted as,
\begin{equation} \label{eq:forward}
 I_{i}^{r'}=\mathcal{M}^{FFSR}(I_{i}^{r}),  \ \ f_{i}=\mathcal{M}^{RIFE}( I_{i}^{r'}),
\end{equation}
where $f_{i}$ is the final feature, $\mathcal{M}^{FFSR}$ and $\mathcal{M}^{RIFE}$ denote the two modules, respectively.

With a training set $\mathbb T = \{(I_{i}^{r}, I_{i}^{h}, P_i)\}, i=1,...,N$, where $I_{i}^{h}$ is the groundtruth high-resolution image and $P_i$ is the person ID label, the network is optimized with two losses computed on two modules, \emph{i.e.},
\begin{equation}\label{wholeLoss}
    \mathcal{L}= \sum_{i=1:N}\mathcal{L}^{FFSR}(I_i^r)+\alpha \mathcal{L}^{RIFE}(I_i^r),
\end{equation}
where $\alpha$ balances the two losses. The following section introduces our network architecture and the implementations of those two loss functions.

\section{Proposed Methods}
\begin{figure*}[t]

  \centering
  \includegraphics[width=0.9\linewidth]{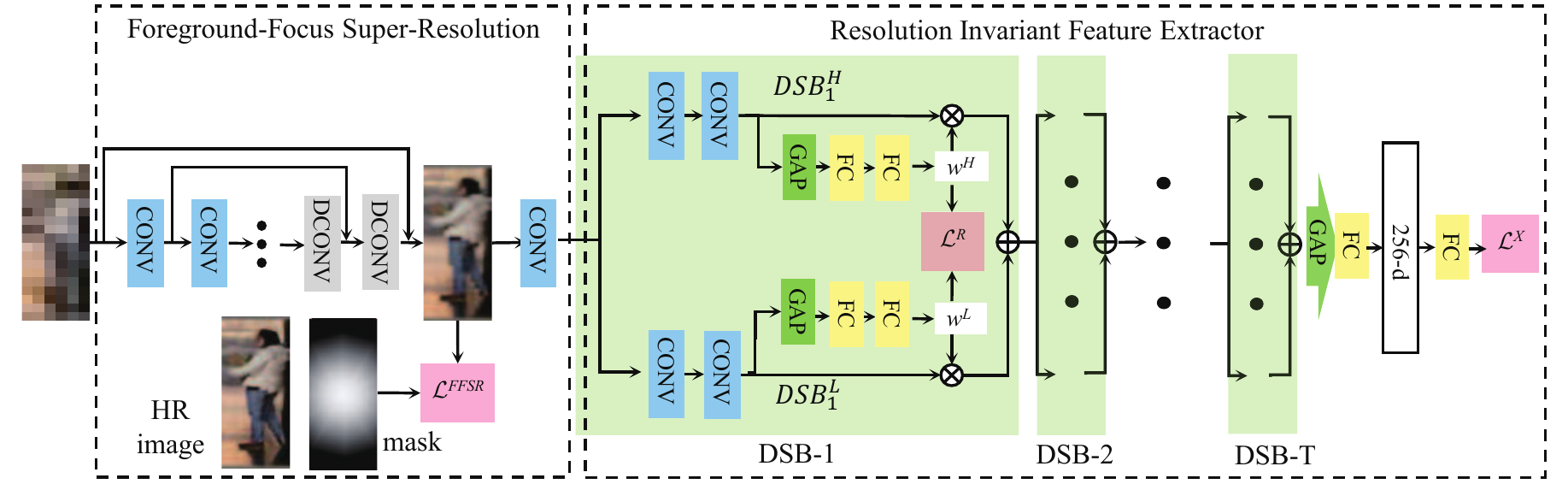}\\
  \caption{The architecture of our network, which consists of two modules: Foreground-Focus Super-Resolution (FFSR) and Resolution-Invariant Feature Extractor (RIFE). FFSR is an auto-encoder with skip connections trained with both person ReID loss and a foreground-focus super resolution loss $\mathcal L^{FFSR}$. RIFE consists of several Dual-Stream Blocks (DSB), each progressively learns resolution invariant features through two CNN streams. Features from two streams are fused with weights learned by a resolution weighting loss $\mathcal L^{R}$. RIFE finally outputs a 256-d feature, which is hence used to compute the cross-entropy loss $\mathcal L^{X}$. }\label{fig:overview}
\end{figure*}

Our network architecture is illustrated in Fig.~\ref{fig:overview}. This section introduces the FFSR and RIFT modules, respectively.

\subsection{Foreground-Focus Super-Resolution}
As the initial stage before feature extraction, FFSR model should be compact and efficient to compute. Additionally, FFSR is expected to work with varied resolutions, \emph{e.g.}, perform super-resolution to low resolution inputs, and preserve original details of high resolution inputs.

Instead of following existing SR models~\cite{kim2016accurate,tai2017image}, we use a light-weight FFSR module illustrated in Fig.~\ref{fig:overview}. FFSR is implemented based on the auto-encoder architecture. The first several convolutional layers down-sample the input with stride width 2. Then, small convolutional kernels with stride width 1 are applied for feature extraction. Following the RED-net~\cite{mao2016image} and U-net~\cite{ronneberger2015u}, we add symmetric skip connections between low and high layers. Skip connects could preserve the visual cues in original images, hence help to enhance the quality of reconstructed images.

Pixel-wised distance Mean Square Error (MSE) is commonly applied for SR model training. Simply minimizing the MSE may not be optimal for person ReID task, because it does not differentiate person foregrounds and backgrounds. Person foregrounds generally provide more valuable cues for person ReID. To recover more visual cues on person foregrounds and depress cluttered backgrounds, we propose the foreground-focus SR loss $\mathcal L_{FFSR}$, \emph{i.e.},
\begin{equation}\label{Masked MSE Loss}
     \mathcal{L}^{FFSR}(I_i^r)=\|M\odot ({I}_{i}^{r'}-I_{i}^{h})\|_{2}^{2},
\end{equation}
where $\odot$ denotes element-wise multiply and $M$ is a mask with the same size of $I_{i}^{r}$.

Our method is compatible with different mask generation strategies. Image segmentation algorithms like \cite{insafutdinov2016deepercut} can be applied to generate binary foreground masks. With a well-trained person bounding box detector, person foregrounds are more likely to appear in the center of bounding boxes. For simplicity, Gaussian kernels can be applied as foreground masks, as illustrated in Fig.~\ref{fig:overview}.

\subsection{Resolution-Invariant Feature Extractor}
Since super-resolution is an ill-posed problem, solely applying FFSR is not strong enough to achieve resolution invariance. We further design RIFE to generate resolution invariant features. As illustrated in Fig.~\ref{fig:difR}, high and low-resolution images convey substantially different amount of visual cues, they should be treated with different feature extractors. RIFE explicitly differentiates high and low resolution images into two feature extraction streams. As shown in Fig.~\ref{fig:overview}, RIFE consists of several Dual-Stream Blocks (DSB), each introduces two feature extraction streams with an identical architecture but different training objectives. The following part first introduces the forward procedure of RIFE, then discusses its training objectives.

In RIFE, each DSB applies two streams of convolutional layers to extract feature maps for high and low-resolution inputs, respectively. For the $t$-th DSB, we denote its two streams as $DSB^L_t$ and $DSB^H_t$, and their generated feature maps as $\textbf{m}_{t}^{L}$ and $\textbf{m}_{t}^{H}$, where superscripts $L$ and $H$ denote the low and high-resolution streams, respectively. $\textbf{m}_{t}^{L}$ and $\textbf{m}_{t}^{H}$ are adaptively fused as the output of the DSB to achieve better robustness to resolution variance. For example, $\textbf{m}_{t}^{L}$ is fused with larger weights for low-resolution images, because $DSB^L_t$ is more suited for feature extraction on low-resolution images. We denote the computation of output feature $\textbf{m}_{t}$ of $t$-th DSB as,
\begin{equation}\label{Eq:DSBoutput}
     \textbf{m}_{t}= w_{t}^L \times \textbf{m}_{t}^{L} + w_{t}^H \times \textbf{m}_{t}^{H},
\end{equation}
where $w_{t}^L$ and $w_{t}^H$ are related to the resolution of input image. For high-resolution images, $w^L$ would be smaller than $w^H$, and vice versa. As shown in Fig.~\ref{fig:overview}, those two weights are predicted with two FC layers based on $\textbf{m}_t^{L}$ and $\textbf{m}_t^{H}$, respectively.

In order to learn $w_t^L$ and $w_t^H$, we introduce the resolution weighting loss $\mathcal L^{R}$ into each DSB. With a training image $I_i^r$, the $\mathcal L^R_t$ for the $t$-th DSB is defined as,
\begin{equation}\label{RW Loss}
\begin{split}
  \mathcal{L}^{R}_{t}(I_i^r)=\|w_t^{L}-(1-r)\|_{2}^{2}+\|w_t^{H}-r\|_{2}^{2},
\end{split}
\end{equation}
where $r$ denotes the resolution of $I_i^r$.

The fused feature map $\textbf{m}_{t}$ is propagated to the next DSB. Stacking multiple DSBs leads to a deep neural network with strong feature learning capability. The output of final DSB is processed with a Global Average Pooling (GAP) layer and a Fully Connected (FC) layer as the final feature $f$. A FC layer is trained on $f$ to predict person ID labels. A cross entropy loss can be computed as the person ReID loss, \emph{i.e.},
\begin{equation}\label{Xent}
    \mathcal{L}^{X}(I_i^r)=CrossEntropy(\operatorname {FC}(f_i), P_i),
\end{equation}
where $P_i$ denotes the person ID label of a training image $I_i^r$.

With $T$ DSBs in total, RIFE is trained with one cross entropy loss and $T$ resolution weighting losses. The RIFE loss on training image $I_i^r$ can be represented as
\begin{equation}\label{updata_FE}
  \mathcal{L}^{RIFE} (I_i^r)=\mathcal L^{X}(I_i^r) +\beta \sum_{t=1:T} \mathcal{L}^{R}_{t}(I_i^r),
\end{equation}
where parameter $\beta$ weights the two losses.

Fusing features with Eq.~\eqref{Eq:DSBoutput} enforces $DSB^L$ and $DSB^H$ to focus on low and high-resolution images during training. For low-resolution images, the back propagated person ReID loss makes more modifications to $DSB^L$ than to $DSB^H$ because of larger $w^{L}$. This mechanism finally learns different parameters for $DSB^L$ and $DSB^H$, respectively and leads to a strong
resolution invariant representation. Implementation details of RIFE will be presented in Sec.~\ref{sec:imp_details}.

\section{Experiment}

\subsection{Datasets}
We evaluate our methods on five datasets, including three existing datasets and two datasets we constructed.

\emph{CAVIAR}~\cite{cheng2011custom} contains 1220 images of 72 identities. Images are captured by one High-Resolution (HR) camera and one Low-Resolution (LR) camera. Among 72 identities, 50 have images from two cameras. Those 50 identities are divided into a LR query set and a HR gallery set.

\emph{MLR-VIPeR} and \emph{MLR-CUHK03} are constructed on \emph{VIPeR} ~\cite{gray2008viewpoint} and \emph{CUHK03} \cite{li2014deepreid} datasets, where both were captured by two cameras.
Following SING~\cite{jiao2018deep}, every image from one camera is down-sampled with a ratio evenly selected from \{$ \frac{1}{2},\frac{1}{3},\frac{1}{4}$\} as the query set. Original images from the other camera are used as the test set. \emph{MLR-VIPeR} has 316 identities for training and testing. \emph{MLR-CUHK03} has 100 identities for testing and 1367 for training, respectively.

\emph{VR-Market1501} and {\emph{VR-MSMT17}} are constructed by us based on \emph{Market1501}~\cite{zheng2015scalable} and \emph{MSMT}~\cite{wei2018person}, respectively. \emph{VR-Market1501} contains 32,217 images of 1,501 people captured by 6 cameras. {\emph{VR-MSMT17}} consists of 126,441 images of 4,101 persons from 15 cameras. All images are down-sampled to make the width within the range of [8, 32) in \emph{VR-Market1501} and [32,128) in \emph{VR-MSMT17}. Hence these two datasets present 24 and 96 different resolutions separately. We keep original divisions of training and testing sets, \emph{i.e.}, 751 and 710 identities for training and testing on \emph{VR-Market1501}, 1,041 and 3,060 for training and testing on \emph{VR-MSMT17}. Compared with existing datasets, \emph{VR-Market1501} and {\emph{VR-MSMT17}} are substantially larger in size and are more challenging, because both query and gallery images show a large range of resolution variance.

\subsection{Implementation Details}\label{sec:imp_details}
Our FFSR module is a 12-layer fully convolutional network. Two convolutional layers with a stride of 2 and two transposed convolutional layers are applied to down-sample and up-sample the feature maps, respectively. We use ResNet50~\cite{he2016deep} as the backbone of RIFE module. Each main block of ResNet50 is modified to be a DSB by duplicating its convolutional layers as $DSB^L$ and $DSB^H$, respectively. Following ResNet50, our RIFE has 4 DSBs. Two FC layers with the output channels of 64 and 1 are used for $w^L$ and $w^H$ prediction.

Our network is trained on PyTorch by Stochastic Gradient Descent (SGD). Training is finished with three steps. 1) We initialize and pre-train the FFSR model on \emph{ImageNet}~\cite{russakovsky2015imagenet} with the MSE loss. Then it is fine-tuned on person ReID training datasets with $\mathcal L^{FFSR}$. 2) We initialize and fine-tune the RIFE module on target dataset with $\mathcal L^{RIFE}$. 3) FFSR and RIFE modules are jointly trained with the loss function in Eq.~\eqref{wholeLoss}. We fix hyperparameters as $\alpha=1, \beta=0.1$ for all datasets. Each step has 60 epoches and the batch size is set as 32. The initial learning rate is set as 0.01 at the first two steps and 0.001 at the final step. The learning rate is reduced ten times after 30 epoches. Input images are resized to $256\times128$ in \emph{VR-market1501} and $384\times128$ in other datasets. The final 256-D feature is used for ReID with Euclidean distance. All of our experiments are implemented with GTX 1080Ti GPU, Intel i7 CPU, and 128GB memory.

\subsection{Ablation Study}

\begin{table}[t]
\begin{center}
\resizebox{0.9\linewidth}{!}{
\begin{tabular}{c|c|c|c|c|c}
\hline
SR model & Mask & JL & Rank-1 & Rank-5 &FLOPs(G)\\
\hline
Bilinear&-&-    &47.6   &66.4 &-\\
SRCNN&- &- &46.7&65.0&2.53\\
VDSR&- &- &48.5&67.7&32.79\\
AE &- & - & 48.1&67.1&3.71\\
FFSR &Gaussian & - & 49.4&67.8&3.71\\
FFSR &Gaussian & $\surd$ &52.8   &69.0&3.71\\
FFSR &Deepercut & $\surd$ &\textbf{52.9}   &\textbf{69.2}&3.71\\
\hline
\end{tabular}
}
\caption{ReID Performance of different SR methods on \emph{VR-MSMT17}. FLOPs shows the SR complexity. JL denotes joint learning with ReID loss.}
\label{tab:2}
\end{center}
\end{table}

\begin{table}[t]
\begin{center}

\resizebox{0.8\linewidth}{!}{
\begin{tabular}{c|c|c|c}
\hline
structure & weight learning  & Rank-1 & Rank-5 \\
\hline
ResNet50&-&47.6   &66.4\\
two ResNet50 &- &49.1&68.2\\
two ResNet50 & $\surd$ &50.4&67.4\\
RIFE & $\surd$&\textbf{53.3}   &\textbf{70.1}\\
\hline
\end{tabular}
}
\caption{Performance of different feature extractors on \emph{VR-MSMT17}}
\label{tab:3}
\end{center}
\end{table}

\begin{figure}[t]
    \centering
     \includegraphics[width=0.95\linewidth]{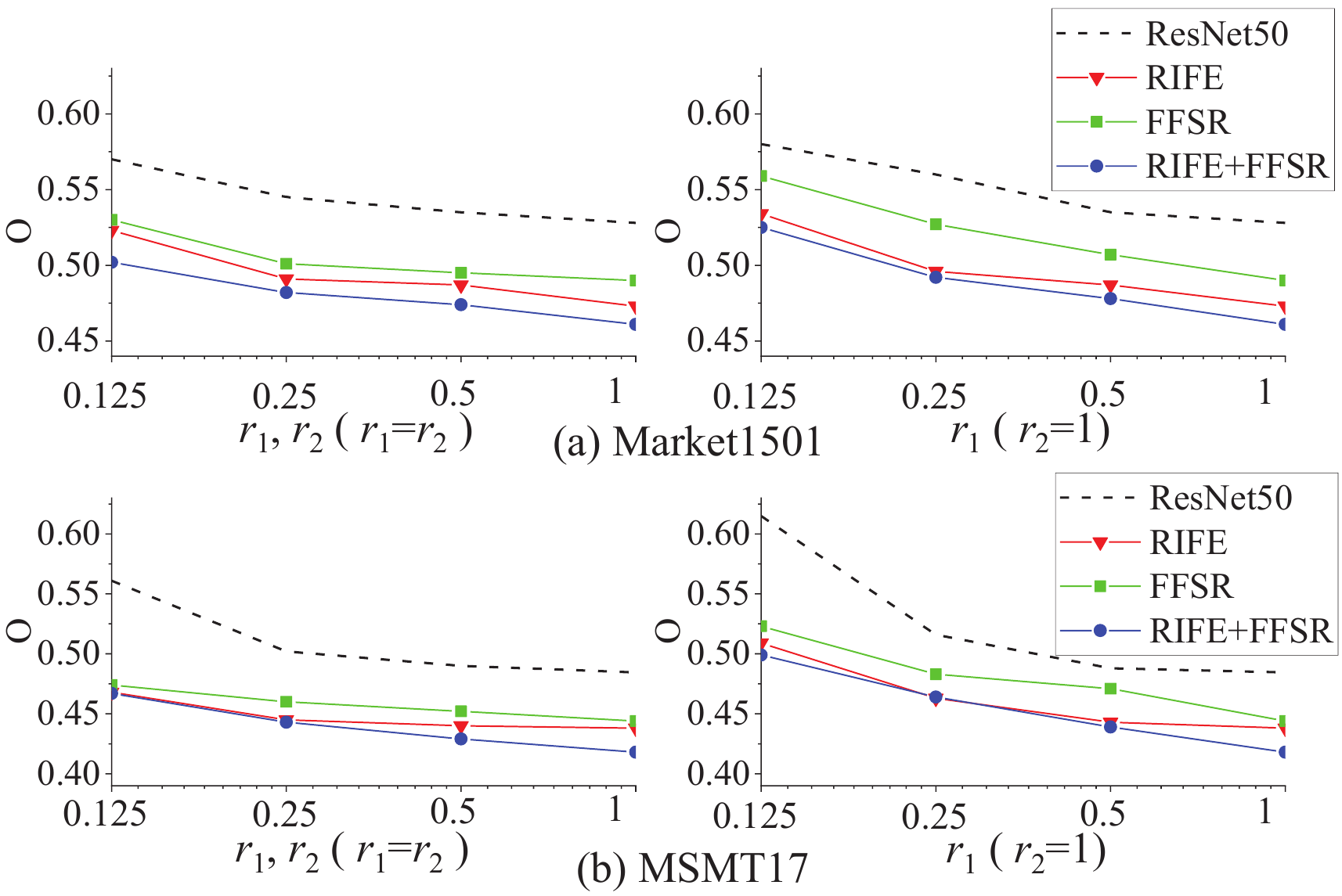}
    \caption{
    Effects of FFSR and RIFT to the object function $\mathcal{O}$ in Eq.~\eqref{eq:obj}. This figure follows the configurations of Fig.~\ref{fig:SwSb}. It is clear that FFSR and RIFE boost the robustness to resolution variance.
    }

    \label{fig:SwSb2}
\end{figure}

\begin{table*}[t]
\begin{center}

\resizebox{0.9\linewidth}{!}{
\begin{tabular}{c|c|c|c|c|c|c|c|c|c|c}
\hline
Dataset   &\multicolumn{2}{c|}{\emph{CAVIAR}}&\multicolumn{2}{c|}{\emph{MLR-VIPeR}}&\multicolumn{2}{c|}{\emph{MLR-CUHK03}}&\multicolumn{2}{c|}{\emph{VR-Market1501}}&\multicolumn{2}{c}{\emph{VR-MSMT17}} \\

\hline
  & Rank-1 & Rank-5& Rank-1 & Rank-5& Rank-1 & Rank-5& Rank-1 & Rank-5& Rank-1 & Rank-5\\
\hline
Densenet121~\cite{Huang2017Densely}& 31.1& 65.5& 31.4& 63.1& 70.8& 91.3&60.0& 78.8&  51.2& 67.4\\
SE-resnet50~\cite{hu2018squeeze}& 30.8& 65.1& 33.5& 63.6&70.8& 92.3&58.2& 78.6& 52.3& 68.9\\
JUDEA~\cite{li2015multi}&20.0   &60.1& 26.0& 55.1& 26.2& 58.0&-&-&-&-\\
SLD$^2$L~\cite{jing2015super}&18.4   &44.8& 20.3& 44.0& -& -&-&-&-&-\\
SDF~\cite{wang2016scale}&14.3   &37.5& 9.52& 38.1& 22.2& 48.0&-&-&-&-\\
SING~\cite{jiao2018deep}&33.5   &72.7 &33.5 &57.0& 67.7& 90.7&60.5 & 81.8&52.1&	68.3\\
CSR-GAN~\cite{wang2018cascaded} &32.3   &70.9 & 37.2 &62.3& 70.7& 92.1& 59.8& 81.3&51.9&	67.5\\
\hline
ResNet50    &29.6 & 64.0   &29.9 & 62.2& 67.4& 91.7& 57.0& 78.7&47.6   &66.4\\
FFSR      &31.1 & 68.7   &40.3 & \textbf{65.3} & 70.5& 92.3& 59.2& 80.1&52.8   &69.0\\
RIFE         &35.7 & \textbf{74.9} &33.9 & 63.6 & 69.7& 91.5& 62.6& 82.4&53.3   &70.1\\
FFSR+RIFE  &$\textbf{36.4}$&72.0&\textbf{41.6}&64.9& \textbf{73.3}& \textbf{92.6}& \textbf{66.9}& \textbf{84.7}&\textbf{55.5}   &\textbf{72.4}\\
\hline

\end{tabular}
}
\caption{Comparison with recent works on five datasets.}
\label{stateofart}
\end{center}
\end{table*}

\vspace {3mm}

\textbf{Validity of FFSR:}
To show the validity of our FFSR model, we fix the feature extraction module as ResNet50 and test different super resolution methods including Bilinear interpolation, SRCNN~\cite{dong2014learning}, VDSR~\cite{kim2016accurate}, as well as variants of our module, \emph{i.e.}, baseline Auto Encoder (AE), FFSR trained with the Gaussian mask and segmented mask by deepercut~\cite{insafutdinov2016deepercut}, as well as training with/without person ReID loss. The experiments are conducted on the large \emph{VR-MSMT17}. We illustrate experimental results in Table~\ref{tab:2}
In Table~\ref{tab:2}, with the Gaussian mask our method outperforms the baseline AE, indicating the validity of emphasizing the person foreground in SR for person ReID. It is also clear that, jointly training with person ReID loss substantially boosts the ReID accuracy. Conveying more accurate foreground locations, segmented mask further outperforms the Gaussian mask. We also compare the computational complexity of FFSR with other super resolution methods. It can be observed that, FFSR introduces marginal computational overhead to the compact SRCNN~\cite{dong2014learning}, but shows substantially better performance, \emph{e.g.}, outperforms SRCNN by 6.2\% in Rank-1 Accuracy. FFSR also substantially outperforms VDSR in the aspects of both accuracy and complexity.

\vspace {3mm}

\noindent\textbf{Validity of RIFE:}
To show the validity RIFE module, we fix the super resolution module as the Bilinear interpolation and compare RIFE with three feature extractors, \emph{i.e.}, a) ResNet50 baseline, b) two ResNet50 with their features fused with equal weight, and c) two ResNet50 with their features fused with learned weights learned with Eq.~\eqref{Eq:DSBoutput}. We summarize the experimental results in Table \ref{tab:3}.
Table \ref{tab:3} shows that increasing the amount of network parameters by fusing two ResNet50 only brings marginal improvements over the one branch ResNet50 baseline. Fusing two ResNet50 with learned weights brings 1.3\% improvements to the Rank-1 Accuracy. Among compared methods, RIFE module achieves the best performance, substantially outperforming baseline by 5.7\% in Rank-1 Accuracy.

\vspace {3mm}

\noindent\textbf{Effect to the Objective Function:} We further show the effect of FFSR and RIFE to the person ReID objective function defined in Eq.~\eqref{eq:obj}. Referring to Fig.~\ref{fig:SwSb}, we illustrate the effect of FFSR and RIFE in  Fig.~\ref{fig:SwSb2}. It is clear that, both FFSR and RIFE decreases the objective $\mathcal{O}$, indicating improved ReID accuracy. It is also clear that, either FFSR and RIFE decrease the slope of the original curves, implying they improve the robustness of a ReID system to resolution variants. Combining FFSR and RIFE brings the best performance. Following part compares our methods with recent works.

\subsection{Comparison with Recent Works}

We compare our method with five recent low-resolution ReID methods including three traditional methods, \emph{i.e.}, JUDEA~\cite{li2015multi}, SLD$^2$L~\cite{jing2015super}, SDF~\cite{wang2016scale}, and two deep learning based methods, \emph{i.e.}, SING~\cite{jiao2018deep}, CSR-GAN~\cite{wang2018cascaded}. Three deep neural networks including ResNet50, Densenet121~\cite{Huang2017Densely}, and SE-resnet50~\cite{hu2018squeeze} are also implemented and compared. We summarize the experimental results on five datasets in Table~\ref{stateofart}, which show the reported performance of JUDEA, SING, SDF, SING, and CSR-GAN on \emph{CAVIAR}, \emph{MLR-VIPeR}, and \emph{MLR-CUHK03}. Performance of compared methods on \emph{VR-Market1501} and \emph{VR-MSMT17} are implemented with the code provided by their authors.

From the comparison we observe that, deep learning based methods substantially outperform the traditional ones. Our method shows promising performance on the first three datasets. On \emph{CAVIAR}, our RIFE module outperforms the recent CSR-GAN by 3.4\% in Rank-1 Accuracy. Combining FFSR and RIFE further boosts the performance, and outperforms CSR-GAN by 4.1\%. On \emph{MLR-VIPeR} and \emph{MLR-CUHK03}, our method outperforms CSR-GAN by 4.4\% and 2.6\% in Rank-1 Accuracy, respectively.

\begin{figure}[t]
    \centering
     \includegraphics[width=0.9\linewidth]{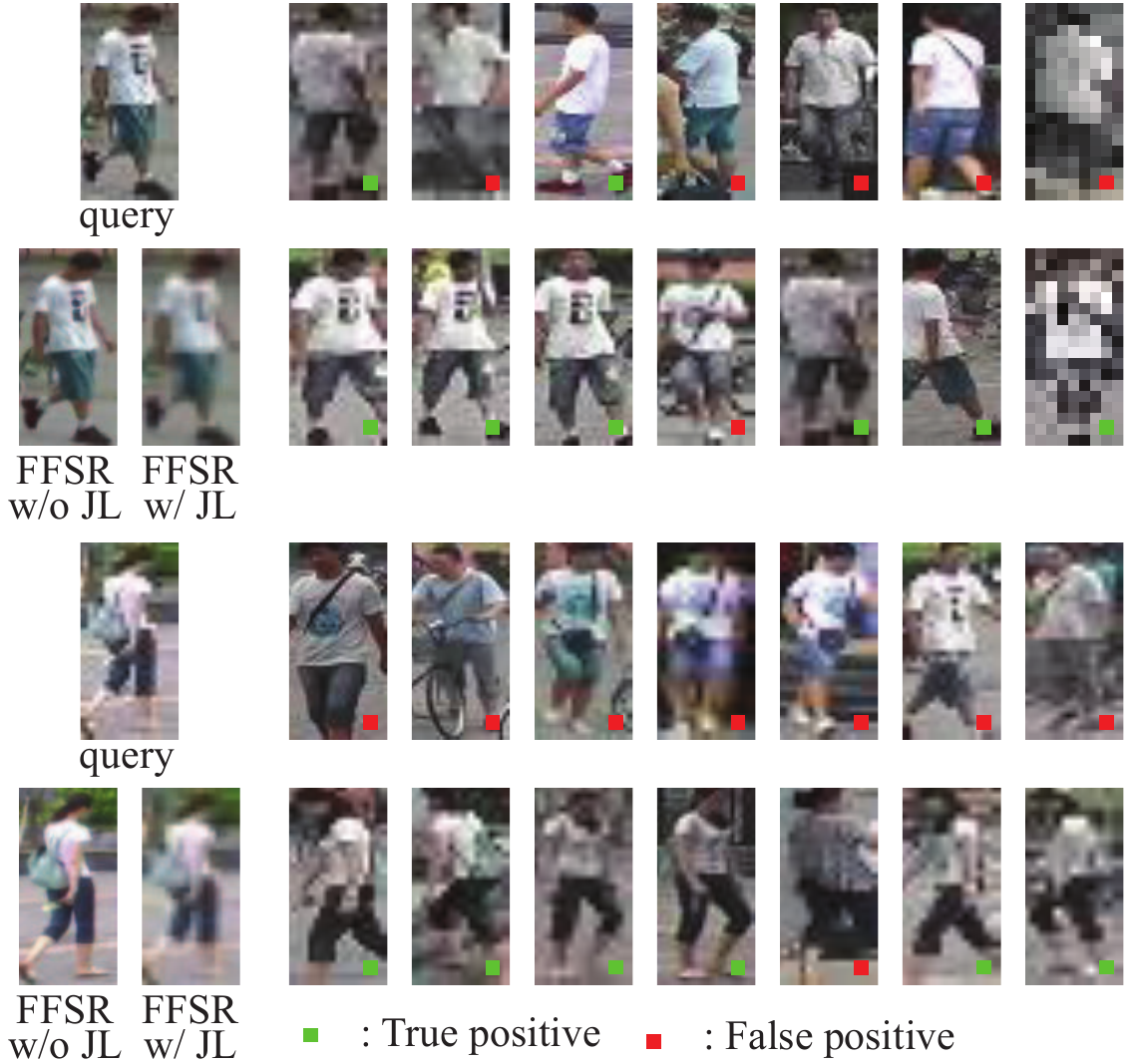}
    \caption{
    Sample results of person ReID and super resolution on \emph{VR-MSMT17}. Our method (second row) outperforms the ResNet50 (first row). It is interesting to observe that, without joint learning with person ReID loss, FFSR module gets a lower ReID accuracy, but produces images with better visual quality.
    }

    \label{fig:results}
\end{figure}

Our method also shows promising performance on \emph{VR-Market1501} and \emph{VR-MSMT17}. Among existing methods that are compared, SING shows the best performance on \emph{VR-Market1501}. FFSR+RIFE outperforms SING by 6.4\%. FFSR and RIFE also achieves the best performance on \emph{VR-MSMT17}, outperforming SE-resnet50 by 3.5\%. It can be observed that, combining FFSR and RIFE commonly leads to the best performance on those five datasets. We show some results of image super resolution and person ReID in Fig.~\ref{fig:results}.

\section{Conclusion}
This paper proposes a deep neural network composed of FFSR and RIFE modules for resolution invariant person re-identification. FFSR upscales the person foreground using a fully convolutional auto-encoder with skip connections learned with a foreground focus loss. RIFE adopts two feature extraction streams weighted by a dual-attention block to learn features for low and high resolution images, respectively. These two complementary modules are jointly trained to optimize the person ReID objective, leading to a strong resolution invariant representation. Extensive experiments on five datasets have shown the validity of introduced components and the promising performance of our methods.

\section*{Acknowledgments}
This work is supported in part by Beijing Natural Science Foundation under Grant No. JQ18012, in part by Natural Science Foundation of China under Grant No. 61620106009, 61572050, 91538111, in part by Peng Cheng Laboratory.
\bibliographystyle{named}
\bibliography{ijcai19}

\end{document}